\documentclass[sn-basic]{sn-jnl}

\usepackage{graphicx}%
\usepackage{multirow}%
\usepackage{amsmath,amssymb,amsfonts}%
\usepackage{amsthm}%
\usepackage{mathrsfs}%
\usepackage[title]{appendix}%
\usepackage{xcolor}%
\usepackage{textcomp}%
\usepackage{manyfoot}%
\usepackage{booktabs}%
\usepackage{algorithm}%
\usepackage{algorithmicx}%
\usepackage{algpseudocode}%
\usepackage{listings}%
\usepackage{lineno}

\begin{document}

% \begin{titlepage}
% \section*{Declarations}
% \bmhead{Funding}
% This work was supported by the Shaanxi Science and Technology Association Youth Talent Support Program(Grant number: 20230115) and the National Natural Science Foundation of China (Grant number: 61802311).
% \bmhead{Conflict of interest}
% The authors have no competing interests to declare that are relevant to the content of this article.

% \bmhead{Ethics approval}
% Not applicable.

% \bmhead{Consent}
% Not applicable.

% \bmhead{Data and Code availability}
% Publicly available data are used. The datasets are available at \url{https://deep-geometry.github.io/abc-dataset/} and \url{https://www.shapenet.org/download/parts}.
% The code is available at \url{https://github.com/Xieyifei1229/EdgeFormer}.

% \bmhead{Author contribution}
% All authors contributed significantly to this work. Yifei Xie experimented with the proposed method. Zhikun Tu conducted comparisons with PBRG, SGLBP, EC-Net, NerVE, and PIE-Net, as well as the ablation studies. Tong Yang processed the experimental results. Yuhe Zhang proposed the main idea. Xinyu Zhou supervised the experiment procedure. All authors read and approved the final manuscript.

% \end{titlepage}

%% \linenumbers
\title{EdgeFormer: Local Patch-based Edge Detection Transformer on Point Clouds}

 \author[1]{\fnm{Yifei} \sur{Xie}}\email{xieyifei@stumail.nwu.edu.cn}\equalcont{These authors contributed equally to this work.}
\author[1]{\fnm{Zhikun} \sur{Tu}}\email{tuzhikun0@gmail.com}\equalcont{These authors contributed equally to this work.}
\author[1]{\fnm{Tong} \sur{Yang}}\email{yangt@stumail.nwu.edu.cn}
\author*[1]{\fnm{Yuhe} \sur{Zhang}}\email{zhangyuhe0601@nwu.edu.cn}
\author*[1]{\fnm{Xinyu} \sur{Zhou}}\email{xiniyuzhou77@163.com}

\affil[1]{\orgdiv{School of Information Science and Technology}, \orgname{Northwest University}, \orgaddress{\street{Xuefu Street}, \city{Xi'an}, \postcode{710127}, \state{Shannxi}, \country{China}}}

\abstract{Edge points on 3D point clouds can clearly convey 3D geometry and surface characteristics, therefore, edge detection is widely used in many vision applications with high industrial and commercial demands. However, the fine-grained edge features are difficult to detect effectively as they are generally densely distributed or exhibit small-scale surface gradients. To address this issue, we present a learning-based edge detection network, named EdgeFormer, which mainly consists of two stages. Based on the observation that spatially neighboring points tend to exhibit high correlation, forming the local underlying surface, we convert the edge detection of the entire point cloud into a point classification based on local patches. Therefore, in the first stage, we construct local patch feature descriptors that describe the local neighborhood around each point. In the second stage, we classify each point by analyzing the local patch feature descriptors generated in the first stage. Due to the conversion of the point cloud into local patches, the proposed method can effectively extract the finer details. The experimental results show that our model demonstrates competitive performance compared to six baselines.}


\maketitle

\section{Introduction}\label{sec1}

3D objects are fundamental and essential components of the real world. Thus the processing of 3D models helps machines to step toward a high-level understanding of the real world. Point clouds contain irregularly distributed points that can highly describe the shapes of 3D objects\cite{Qi2017Pointnet, Guo2021PCT, Tang2022Contrastive, Feng2021Relation}, and thus are widely used in various 3D model processing and understanding research, including 3D model classification\cite{Qi2017Pointnet, Guo2021PCT}, semantic segmentation\cite{Tang2022Contrastive}, and object detection\cite{Feng2021Relation}. Edge points of point cloud carry necessary information about the entire shape, local structure, and silhouette of smooth surfaces. Thus edge detection is necessary and important to several point cloud processing tasks, such as point cloud simplification\cite{Wang2018Saliency}, non-realistic rendering\cite{Zhang2014Splatting}, 3D reconstruction\cite{Metzer2021Self}, etc.

Edges often appear in the intersection region of at least two surfaces, meaning that surface gradient is an important indicator of edges. However, due to the scattered and irregular sampling pattern of points on the point clouds, it is difficult to measure the scale of the surface gradient, neither globally nor locally, especially for point clouds containing sharp, smooth, and fine-grained edges.

Existing non-learning-based methods mainly leverage geometric features\cite{Pauly2003Multi, Xia2017AFast} and local neighborhood information of point clouds\cite{Guo2022SGLBP}. Additionally, some other approaches utilize surface fitting strategies \cite{Daniels2008Spline} or transform the point cloud into images or meshes. Compared with non-learning methods, learning-based methods are more effective in terms of both fitting and efficiency ability. However, the existing learning-based methods, such as EC-Net\cite{Yu2018ECNet} fitting edges on surface patches, PIE-Net\cite{Wang2020PieNet} inferring parametric curves, and NerVE\cite{Zhu_2023_NerVE} separating points in voxels, often take either the entire point cloud or a large surface patch (\textit{e.g.}, containing 1024 points) as input, rendering it challenging for these methods to handle finer detailed structures effectively.

To tackle the above-mentioned challenge, we design an easy-implemented network, named EdgeFormer, to detect edge points on point clouds. Specifically, to achieve fine-grained edge identification, all points on the point cloud should be analyzed to distinguish edges from non-edges. Therefore, in this work, the surface patches are formed by each point $p$ and its ball neighborhood. Thus a point cloud with $N$ points can thereby form $N$ training samples.

% , resulting in a large decrease of training models, \textit{i.e.}, only \colorbox{yellow}{50} point clouds are sufficient for achieving a superior performance.

\begin{figure}[htbp]%调节图片位置，h：浮动；t：顶部；b:底部；p：当前位置
	\centering
	\includegraphics[width=1\linewidth]{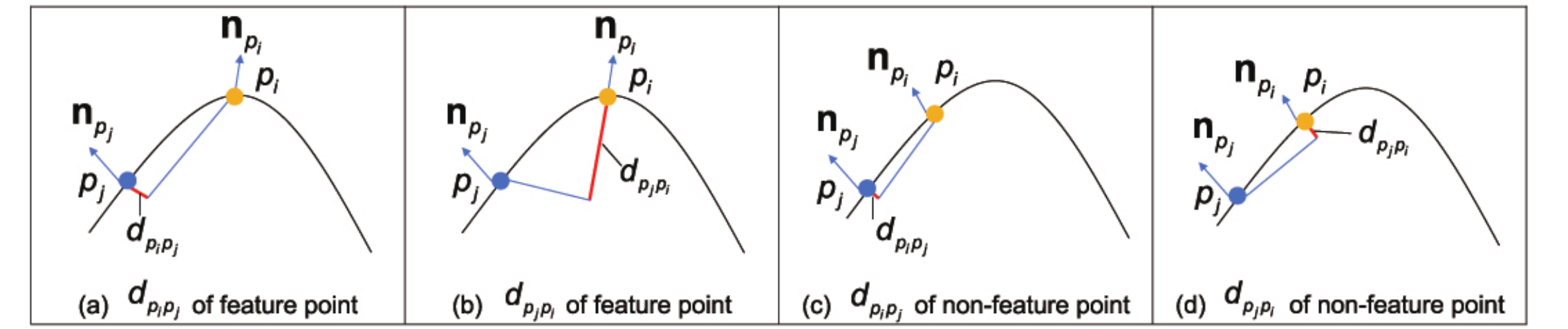}
	\caption{Schematic diagram of projection distance between feature point and non-feature point}
	\label{Figure 1}
\end{figure}

As for the definition of edges, we hold a point that edges generally exhibit large- or small-scale surface gradients, resulting in relatively large angles between the normals of the target point and its neighbors. Correspondingly, for edge point $p_i$, the difference between $d_{{p_i}{p_j}}$ and $d_{{p_j}{p_i}}$ is large, whereas for non-edge point $p_i$, the difference between $d_{{p_i}{p_j}}$ and $d_{{p_j}{p_i}}$ is relatively small, as shown in Figure~\ref{Figure 1}, $d_{|\dot|}$ is the projection distance. Therefore, we employ $d_{{p_i}{p_j}}$ and $d_{{p_j}{p_i}}$ to approximate the local surface gradient at a target point $p_i$, thereby forming the local patch feature descriptor of $p_i$.

Given that the probability of being an edge point depends on the number and distribution of the non-edge points and edge points in the neighborhood (the same goes for non-edge points), the relationship between $p_i$ and its neighbors $p_j (p_j \in kNN(p_i) )$ is highly beneficial for identifying the edge points. Hence, we use Transformer to encode the relationship between $d_{{p_i}{p_j}}$ and $d_{{p_j}{p_i}}$, thereby distinguishing edges from non-edges.

We experimentally compare the performance of EdgeFormer with six methods on 6000 models from the ABC dataset, including BE\cite{Rusu20113D}, PBRG\cite{Zhang2016Statistical}, SGLBP\cite{Guo2022SGLBP}, which are non-learning based methods, and EC-Net\cite{Yu2018ECNet}, PIE-Net\cite{Wang2020PieNet}, NerVE\cite{Zhu_2023_NerVE}, which are deep-learning-based methods. The evaluation metrics contain Hausdorff distance, IoU, MCC, Precision and Recall. Furthermore, we also compared the baselines and our method on the PartNet dataset\cite{Mo_2019_PartNet} to further validate their generalization capabilities.

Both quantitative and qualitative evaluations demonstrate that our method outperforms the selected baseline methods across the six evaluation metrics. In addition, from a qualitative standpoint, our approach demonstrates the ability to detect sharp, smooth, and fine-grained features on point clouds.

Our main contributions can be summarized as follows:

\begin{itemize}

\item We propose a novel network named EdgeFormer that can effectively detect both sharp and smooth edges, as well as detailed structures in point clouds.

\item Our method is first to utilize the Transformer to detect edges in point clouds. By using $k$NN to form the surface patch, the computational complexity of the Transformer is highly reduced.

\item The presentation of $d_{{p_i}{p_j}}$ and $d_{{p_j}{p_i}}$ for encoding the local surface patch serves as the input to EdgeFormer, improving the performance of edge detection.

\end{itemize}

\section{Related Work}
\subsection{Non-learning-based methods for feature detection}
Research on non-learning-based methods includes methods based on local geometric features and neighborhood information of point clouds, methods based on surface fitting, and methods based on images or meshes obtained from original point clouds.

Methods based on local geometric features and neighborhood information focus on analyzing the geometric properties and interrelationships of points and their local neighborhoods. Through this local information, important structural features can be identified, such as edges, corners, planes, \textit{etc.}. Mérigot \textit{et al.}\cite{Mérigot2011Voronoi} used the covariance matrices of Voronoi cells to extract curvature information, sharp features, and normal directions from point clouds. Pauly \textit{et al.}\cite{Pauly2003Multi} discreted surface analysis at multiple scales, and then used a minimum spanning graph to extract feature lines. PBRG\cite{Zhang2016Statistical} also analyzed the normals of points on the surface to identify the features. Xia\textit{et al.}\cite{Xia2017AFast} proposed a new edge detector based on the geometric center and detected edge candidates by analyzing eigenvalues. Demarsin \textit{et al.}\cite{Demarsin2007Detection} utilized the Principal Component Analysis (PCA) and cluster analysis to distinguish feature points. Bazazian \textit{et al.}\cite{Bazazian2015Fast} detected sharp edge features by analyzing the eigenvalues of the covariance matrix defined by the k-nearest neighbors of each point. Gumhold \textit{et al.}\cite{Gumhold2001Feature} used covariance analysis to analyze the distribution of points for feature extraction. Weber \textit{et al.}\cite{Weber2010Sharp} used Gauss map clustering on local neighborhoods for sharp feature detection. Chen \textit{et al.}\cite{Chen2022Multiscale} defined a statistical metric to detect all potential feature points, and then used an anisotropic contracting scheme to force feature points lying on the underlying real feature lines. SGLBP\cite{Guo2022SGLBP} also analyzed neighborhood information for detecting edge features based on Local Binary Patterns.

Another category of methods is surface fitting, which often fits a surface in a local area to extract features. Daniels \textit{et al.}\cite{Daniels2007Robust} and Daniels II \textit{et al.}\cite{Daniels2008Spline} used robust moving least squares to fit surfaces and projected each point to its nearest intersection between the surfaces. Multi-level implicit surface fitting was employed by Ohtake \textit{et al.}\cite{Ohtake2004Ridge}.

The other methods first preprocess the point clouds to obtain images\cite{Lin2015Line} or meshes\cite{Stylianou2004Crest, Hildebrandt2005Smoot, Gao2019Extraction}, and then calculate based on these forms, however, they are different from the method in this work.

\subsection{Learning-based methods for feature detection}
With the rapid development of machine learning, many methods have been proposed, such as Support Vector Machines(SVM)\cite{Wang2019Development}, or random forest\cite{Hackel2016Contour}.

Likewise, neural networks have gained significant attention. Yu \textit{et al}.\cite{Yu2018ECNet} proposed to realize edge detection by fitting edges in the upsampling work. PIE-Net\cite{Wang2020PieNet} designed a network that is about parameter inference of feature edges over a 3D point cloud. Furthermore, Loizou \textit{et al}.\cite{Loizou2020Learning} used a graph convolutional network to learn part boundaries, but the effect for sparse or tiny features is still not satisfactory. DEF\cite{Matveev2022DEF} was introduced to regress a scalar field instead of feature classification. Edge detection and semantic segmentation were jointly performed in JSENet\cite{Hu2020JSENet}. Similarly, Bazazian and Parés\cite{Bazazian2021EDC} proposed EDCNet which utilizes both edge labels and segmentation labels to train the network. Both PCEDNet\cite{Himeur2021PCEDNet} and BoundED\cite{Bode2022BoundED} build descriptors that can distinguish feature points and non-feature points to classify points in the point cloud. Feng \textit{et al}.\cite{Feng2023Deep} encode the features using the structure of U-net and detect local information using a convolution operation. NEF\cite{Ye_2023_NEF} utilizes a neural radiance field to regress the feature curves of objects from multi-view images. Zhu \textit{et al}.\cite{Zhu_2023_NerVE} proposed a novel neural volumetric edge representation called NerVE. NerVE captures three corresponding attributes for each voxel: edge occupancy, edge orientation, and edge point location. These attributes are then transformed into a universal piece-wise linear (PWL) curve representation, and finally, precise parametric curves are obtained through spline fitting algorithms.

\subsection{Transformers on Point Clouds}
Inspired by the success of Transformer\cite{Vaswani2017Attention} in Natural Language Processing (NLP), recently researchers also extended the Transformer structure for various point cloud tasks like point cloud classification\cite{Lu20223DCTN, Guo2021PCT, Zhao2021Point}, point cloud upsampling\cite{Du2022Pint}, point cloud segmentation\cite{Guo2021PCT, Zhao2021Point} and point cloud completion\cite{Yu2021PoinTr}. Guo \textit{et al}.\cite{Guo2021PCT} proposed PCT, which is a Transformer-based point cloud learning framework. PCT leverages the inherent order invariance of the Transformer to avoid defining the order of point cloud data and using the attention mechanism for feature learning, making it more suitable for point cloud feature learning. Du \textit{et al}.\cite{Du2022Pint} introduced a cascaded refinement network for point cloud upsampling and designed a feature extraction module based on transformers to learn global and local shape contexts. Yu \textit{et al}.\cite{Yu2021PoinTr} also employed a Transformer encoder-decoder architecture to predict missing parts of the point cloud. Different from the above methods, we applied the Transformer on point cloud edge detection.

\section{Method}

In this section, we elaborate technical details of our method. The proposed approach mainly consists of two stages. The first stage generates the descriptors of the local patches that are formed by the neighborhood of each point on the point clouds, and the second stage utilizes the EdgeFormer to analyze the descriptors of local patches and classify the points into edge points and non-edge points.

\subsection{Local Patch Encoding}

Given a point cloud $P = \{ {p_i}\} ,1 \le i \le N$, $N$ denotes the number of points on $P$. The normal vectors of the points are estimated using the Re-Compute Vertex Normals method \cite{Grit1998Computing, Nelson1999Weights} implemented in MeshLab since this step is not the core of the proposed method. Subsequently, the $k$NN ($k$-Nearest Neighbors) algorithm is applied to each point $p_i$ to obtain its nearest set of points $\{ {p_j}\} _{j = 1}^K$, forming the local patch.

Given the local patches, the projection distances $d_{p_ip_j}$ and $d_{p_jp_i}$ are calculated using Equation~\eqref{Equation 1} and Equation~\eqref{Equation 2}:

\begin{equation}
    \label{Equation 1}
    {d_{{p_i}{p_j}}} = \left\{ {\begin{array}{*{20}{c}}
    {|{p_i} \cdot {{\bf{n}}_{{p_j}}} - {p_j} \cdot {{\bf{n}}_{{p_j}}}|}&{{\rm{if }}|{p_i} \cdot {{\bf{n}}_{{p_j}}} - {p_j} \cdot {{\bf{n}}_{{p_j}}}| \ge {{10}^{ - 6}}}\\
    0&{otherwise}
    \end{array}} \right\},1 \le i \le N,1 \le j \le K
\end{equation}

\begin{equation}
    \label{Equation 2}
    {d_{{p_j}{p_i}}} = \left\{ {\begin{array}{*{20}{c}}
    {|{p_j} \cdot {{\bf{n}}_{{p_i}}} - {p_i} \cdot {{\bf{n}}_{{p_i}}}|}&{{\rm{if}}|{p_j} \cdot {{\bf{n}}_{{p_i}}} - {p_i} \cdot {{\bf{n}}_{{p_i}}}| \ge {{10}^{ - 6}}}\\
    0&{otherwise}
    \end{array}} \right\},1 \le i \le N,1 \le j \le K
\end{equation}
where ${d_{{p_i}{p_j}}}$ represents the projection distance of $p_i$ on the normal of $p_j$, and ${d_{{p_j}{p_i}}}$ is the projection distance between of $p_j$ on the normal of $p_i$. $K$ is the number of points in the local patch, which is set to 20, in this work. Ultimately, two sets of local patch feature descriptors are obtained, designated as ${D_1} = \{ {d_{{p_i}{p_j}}}\} _{i = 1,j = 1}^{N, K}$ and ${D_2} = \{ {d_{{p_j}{p_i}}}\} _{i = 1,j = 1}^{N, K}$.

\subsection{EdgeFormer Architecture}

\begin{figure}[htbp]
	\centering
	\includegraphics[width=1\linewidth]{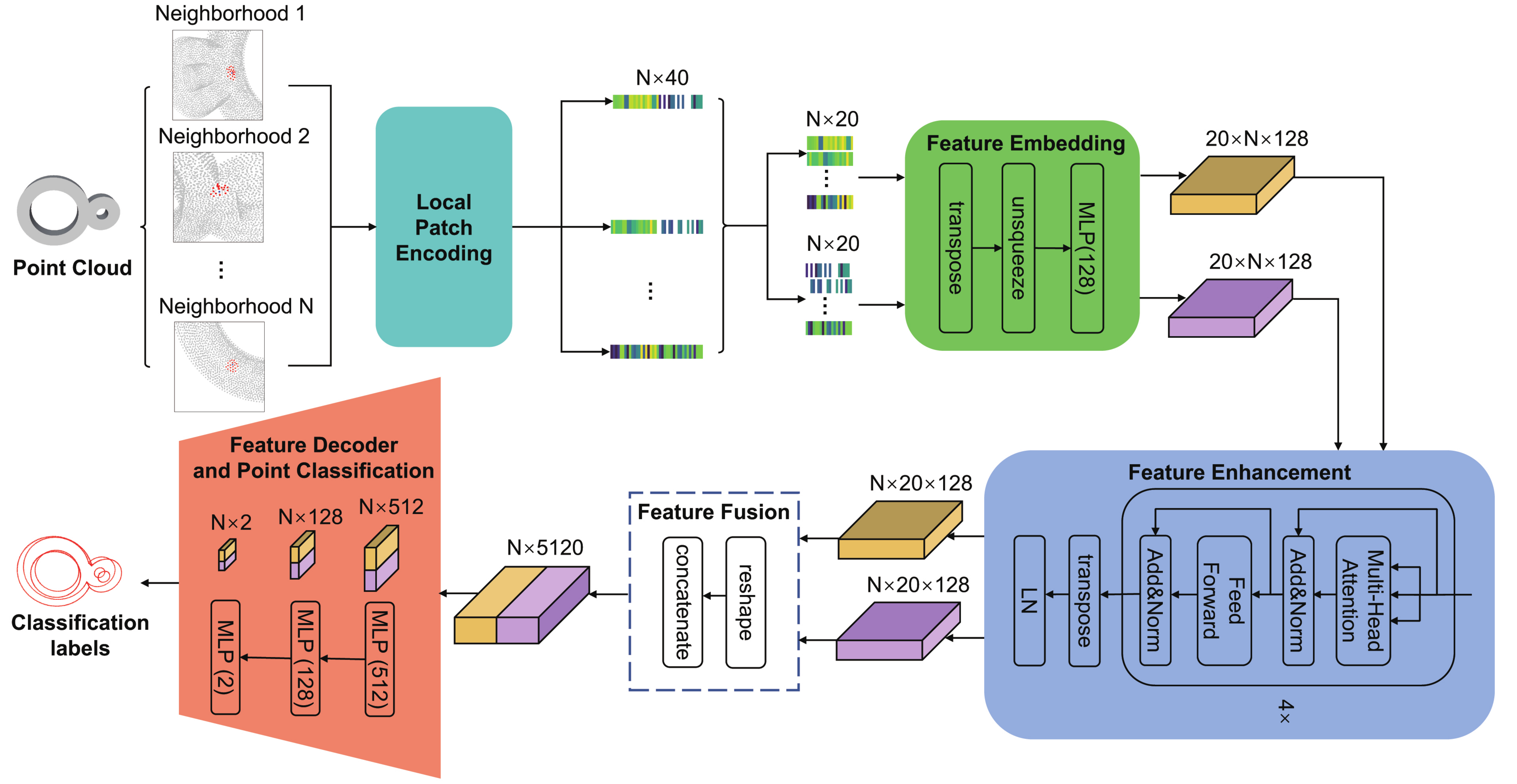}
	\caption{The overall architecture of EdgeFormer}
	\label{Figure 2}
\end{figure}

EdgeFormer architecture is shown in Figure~\ref{Figure 2}, including four modules: feature embedding, feature enhancement, feature fusion, and feature decoder and point classification. These modules work collaboratively to fully perceive and explore both global and local features within local patches, thereby achieving accurate point cloud edge detection.

\textbf{Feature Embedding.} To guide the Transformer in modeling the local patch context, we design the feature embedding module to embed feature descriptors within the local patch. Specifically, we input two sets of local patch feature descriptors $D_1$ and $D_2$ with dimensions $N \times K$ into transpose, unsqueeze, and MLP in sequence. Transpose and unsqueeze are responsible for dimension transformation and increase of the two sets of local patch feature descriptors, while MLP performs dimension mapping. The configuration of MLP is (1,128), and the activation function is ReLU. The outputs are ${F_{emd1}}$ and ${F_{emd2}}$, both of which have dimensions of $K \times N \times 128$, as shown in Equation~\eqref{Equation 3} and Equation~\eqref{Equation 4}.

\begin{equation}
    \label{Equation 3}
    {F_{emd}}_1 = {\rm{ML}}{{\rm{P}}_{128}}({\rm{unsqueeze}}({\rm{transpose}}({D_1},1,2)),3))
\end{equation}

\begin{equation}
    \label{Equation 4}
    {F_{emd}}_2 = {\rm{ML}}{{\rm{P}}_{128}}({\rm{unsqueeze}}({\rm{transpose}}({D_2},1,2)),3))
\end{equation}

\textbf{Feature Enhancement.} To extract enhanced information for the local patch information of points, we first utilize the Transformer encoder\cite{Vaswani2017Attention} to model the relationships of neighborhood points and obtain global representations, followed by LayerNorm regularization. The final outputs are the local patch feature map ${F_{enhance1}}$ and ${F_{enhance2}}$ (defined in Equation~\eqref{Equation 5} and Equation~\eqref{Equation 6}), with each having dimensions of $N \times K \times 128$. We set the number of encoder layers to 4 and the head number of multi-head attention is 8.

\begin{equation}
    \label{Equation 5}
    F_{enhance1} = {\rm{LayerNorm(transpose}}({\rm{Encoder}}({F_{emd}}_1),1,2))
\end{equation}

\begin{equation}
    \label{Equation 6}
    F_{enhance2} = {\rm{LayerNorm(transpose}}({\rm{Encoder}}({F_{emd}}_2),1,2))
\end{equation}

 \textbf{Feature Fusion.} In the feature fusion module, the last two dimensions of $F_{enhance1}$ and $F_{enhance2}$ are transformed into a single dimension and concatenated. This process avoids information loss, resulting in a fused local patch shape descriptor $F_{fuse}$ defined in Equation~\eqref{Equation 7}.

\begin{equation}
    \label{Equation 7}
    F_{fuse} = {\rm{cat}}({\rm{reshape}}(F_{enhance1}),{\rm{reshape}}(F_{enhance2})))
\end{equation}

\textbf{Feature decoder and Point Classification.} The feature decoder and point classification module transform fused descriptors into vectors representing category probabilities. Specifically, this module feeds $F_{fuse}$ into an MLP with two hidden layers, which have 512 and 128 nodes, respectively. To enhance the model's ability to express non-linear relationships, each hidden layer includes a ReLU activation function and a Batch Normalization (BatchNorm) module. To avoid overfitting, a dropout layer with drop probability of 0.5 is incorporated between the hidden layers. The process concludes with an output layer of size 2 in the MLP, followed by another application of BatchNorm, resulting in the vector of category probabilities $Output$ in Equation~\eqref{Equation 8}.

\begin{equation}
    \label{Equation 8}
    Output = {\rm{ML}}{{\rm{P}}_2}({\rm{ML}}{{\rm{P}}_{128}}({\rm{Dropou}}{{\rm{t}}_{0.5}}({\rm{ML}}{{\rm{P}}_{512}}(F_{fuse}))))
\end{equation}

\subsection{Implementation Details}
To deal with the unbalanced classes, we set the ratio of feature points and non-feature points to 1:1. We use 50 models for training, 200 models for validation, and 6000 models for testing.

The generation of local descriptors is implemented using Scipy and Numpy libraries. The network was developed on the PyTorch platform, utilizing an NVIDIA RTX 3080Ti GPU. Additionally, the network was trained with the Adam Optimizer. The initial learning rate was set to 1e-6, with milestones at [75, 150] and gamma set to 0.1 for learning rate adjustments. We trained the network for 200 epochs and set the batch size to 64. The cross-entropy loss function was employed as the loss function.

\section{Experiments and Results}
In this section, we first evaluate the performance and generalization capability of EdgeFormer. Subsequently, we assess its effectiveness and robustness through experiments on noisy models and down-sampled models. Finally, we conduct ablation studies to validate the design of the proposed architecture and compare the runtime of different models.

\subsection{Datasets}
\subsubsection{ABC Dataset}
\begin{figure}[htbp]%调节图片位置，h：浮动；t：顶部；b:底部；p：当前位置
	\centering
	\includegraphics[width=1\linewidth]{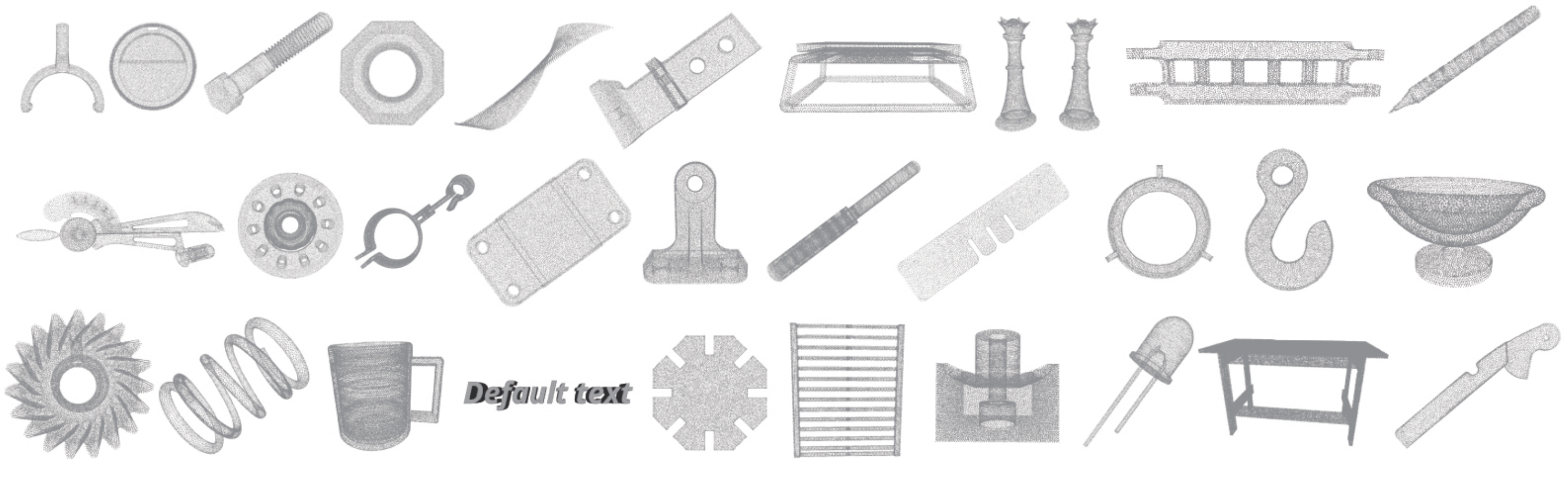}
	\caption{Some examples of ABC dataset}
	\label{Figure 3}
\end{figure}

The ABC dataset\cite{Koch2019ABC} is a large, open-source CAD model dataset that includes a series of mechanical component models with well-defined surfaces and sharp edges, making it suitable for edge detection tasks. More specifically, it comprises models, parametric curve annotations, and other data. The parametric curve annotations are stored in YAML files. Our experiments utilize chunk 000 from the ABC dataset. Some examples are shown in Figure~\ref{Figure 3}.

The YAML files in the ABC dataset provide various information on the parametric curves of point clouds, primarily categorized into two types: feature lines and feature surfaces. Feature lines are further subdivided into lines, circles, ellipses, B-spline curves, and other types, each containing details such as curve names, locations, and vertex indices. To construct the ground truth dataset for point cloud edge features, the following steps are employed: firstly, YAML files are loaded to retrieve configuration information; then, feature lines are read, and if the 'sharp' attribute of a curve is marked as 'true,' the list of vertex indices for that curve is stored in a specific list. These indices are utilized to extract the corresponding points from the original point cloud and preserve them as Ground Truth.

\subsubsection{PartNet Dataset}
\begin{figure}[htbp]%调节图片位置，h：浮动；t：顶部；b:底部；p：当前位置
\centering
\includegraphics[width=1\linewidth]{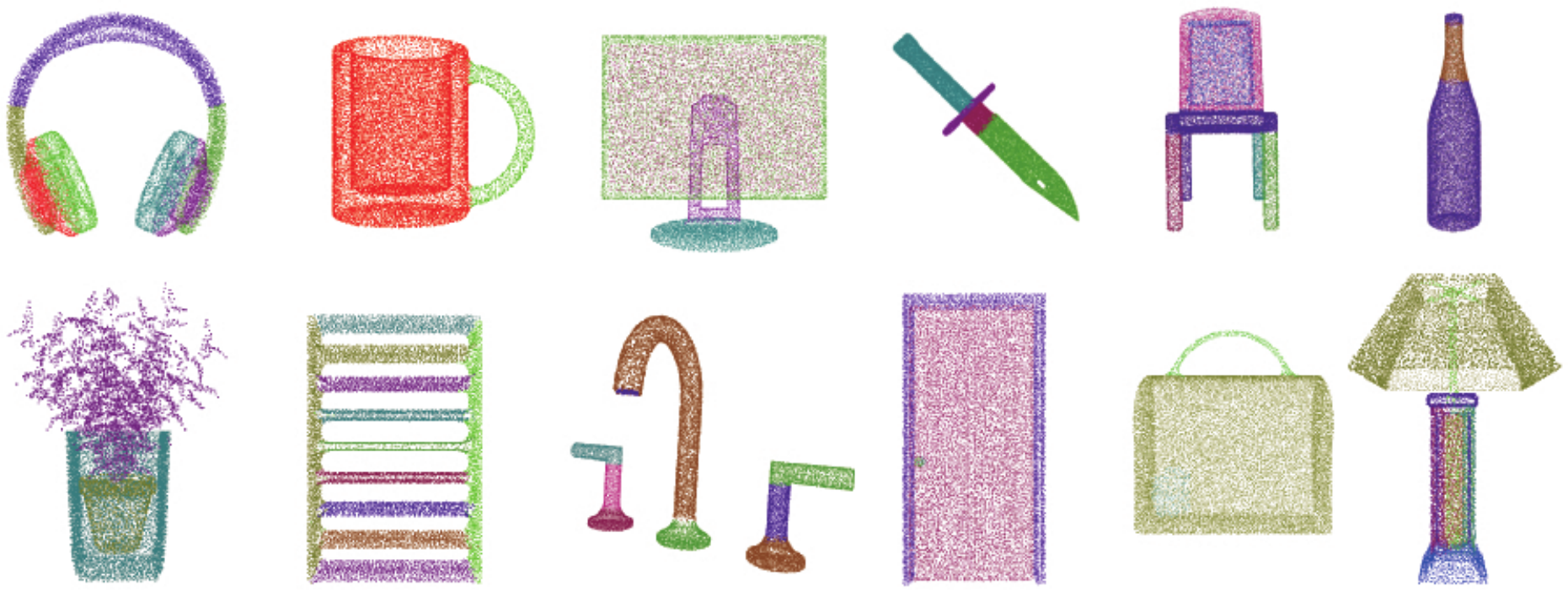}
\caption{Some examples of PartNet dataset}
\label{Figure 4}
\end{figure}

The PartNet dataset\cite{Mo_2019_PartNet} is a consistent, large-scale 3D object dataset focused on providing a fine-grained and hierarchical part-level understanding of 3D objects, commonly used for semantic segmentation. It comprises 573585 part instances from 26671 3D models covering 24 object categories. This dataset supports and acts as a catalyst for numerous tasks such as shape analysis, dynamic 3D scene modeling and simulation, and affordance analysis. In our experiments, we employed point cloud models that are equipped with normal vectors and sampled at 10000 points. Some examples are presented in Figure~\ref{Figure 4}.

\subsection{Baseline methods}
We compare our methods with six baseline methods which include BE\cite{Rusu20113D},  PBRG\cite{Zhang2016Statistical}, SGLBP\cite{Guo2022SGLBP}, EC-Net\cite{Yu2018ECNet}, PIE-Net\cite{Wang2020PieNet}, and NerVE\cite{Zhu_2023_NerVE}. The former three baseline methods are non-learning-based methods and the latter three are based on deep learning.

BE is implemented by PCL, utilizing an angle threshold to estimate whether a set of points is lying on the surface boundaries. PBRG uses Poisson distribution which is a statistical model to detect feature points. SGLBP encodes the topology of the neighborhoods of the points. EC-Net performs edge detection during 3D reconstruction. PIE-Net proposes a new perspective that utilizes deep networks to estimate parametric edge curves. NerVE performs voxel analyses on point clouds and subsequently learns features based on neural volume edge representations.

All the baseline methods utilize the default parameters recommended in the references. The codes of PBRG, SGLBP, EC-Net, PIE-Net, and NerVE are provided by the authors. BE is implemented using PCL. Furthermore, PIE-Net uses a downsampled point cloud containing 8096 points as input, whereas BE, PBRG, SGLBP, EC-Net, and NerVE utilize the original point clouds without any sampling. The training point clouds for EC-Net are 36, for PIENet are 32, and NerVE requires 1892 point clouds to train the network.

\subsection{Evaluation Metrics}

Since the outputs of EC-Net and PIE-Net are fitted points, not the original feature points on point clouds, we normalized the outputs of baseline methods and Ground Truth. Furthermore, the corresponding point pairs are calculated by ICP\cite{Paul_1992_Method}. To evaluate the edge detection results more accurately, we employed various evaluation metrics, including $Hausdorff distance$, $IoU$, $MCC$, $Precision$, $Recall$. $Pred$ is the edges predicted by EdgeFormer and baselines, and $GT$ is the Ground Truth.

Hausdorff distance is calculated using Equation~\eqref{Equation 10}\cite{Moscoso2019SHREC}:

\begin{equation}
    \label{Equation 9}
    {d_{Hausdorff}}(Pred,GT) = ma{x_{({p_a} \in Pred)}}mi{n_{({p_b} \in GT)}}d({p_a},{p_b})
\end{equation}

\noindent where $d$ is the Euclidean distance. ${p_a}$ and ${p_b}$ are corresponding point pairs in $Pred$ and $GT$. And then, the widely known Hausdorff distance is calculated by Equation~\eqref{Equation 10}:

\begin{equation}
    \label{Equation 10}
    max\{ {d_{Hausdorff}}(Pred, GT),{d_{Hausdorff}}(GT, Pred)\}
\end{equation}

Furthermore, we also use True Positives (TP), False Positives (FP), True Negatives (TN), and False Negatives (FN) to calculate Precision, Recall, MCC, IoU, and the corresponding point pairs achieved by ICP are regarded as TP. They are defined as:

\begin{equation}
    \label{Equation 11}
    Precision = \frac{{TP}}{{TP + FP}}
\end{equation}

\begin{equation}
    \label{Equation 12}
    Recall = \frac{{TP}}{{TP + FN}}
\end{equation}

\begin{equation}
    \label{Equation 13}
    MCC = \frac{{TP \times TN - FP \times FN}}{{\sqrt {(TP + FP)(TP + FN)(TN + FP)(TN + FN)} }}
\end{equation}

\begin{equation}
    \label{Equation 14}
    IoU = \frac{{TP}}{{TP + FP + FN}}
\end{equation}

\subsection{Quantitative Comparison}
In this section, we present a quantitative evaluation of the features detected by EdgeFormer and baseline methods, using 6000 models from the ABC dataset. Table~\ref{Table 1} reports the results which contain six evaluation metrics.

\begin{table*}[htbp]
  \centering
  %\fontsize{10}{12}\selectfont
  \begin{threeparttable}
  \caption{Comparison of different methods on the ABC dataset. The best scores are in bold}
  \label{Table 1}
  
    \begin{tabular}{lcccccc}
\hline
Method & $Hausdorff \downarrow$ & $IoU \uparrow$ & $MCC\uparrow$ & $Precision\uparrow$ & $Recall\uparrow$
\\ \hline
BE\cite{Rusu20113D} & 0.486 & 0.165 & 0.182 & 0.271 & 0.469 \\
PBRG\cite{Zhang2016Statistical} & 0.316 & 0.206 & 0.261 & 0.349 & 0.408 \\
SGLBP\cite{Guo2022SGLBP} & 0.252 & 0.255 & 0.275 & 0.350 & 0.543 \\
EC-Net\cite{Yu2018ECNet} & 0.248 & 0.066 & -0.084 & 0.081 & 0.421\\
PIE-Net\cite{Wang2020PieNet} & 0.354 & 0.070 & 0.161 & 0.309 & 0.082\\
NerVE\cite{Zhu_2023_NerVE} & 0.252 & 0.388 & 0.541 & \textbf{0.894} & 0.401\\
EdgeFormer(ours) & \textbf{0.115} & \textbf{0.839} & \textbf{0.885} & 0.890 & \textbf{0.922}\\ \hline
\end{tabular}
    \end{threeparttable}
\end{table*}

In the various metrics evaluated, BE does not exhibit significant advantages. PBRG and SGLBP show similar performance on the evaluation metrics. However, SGLBP slightly outperforms PBRG across all metrics, particularly regarding IoU and Recall. Owing to EC-Net identifying fitted points instead of the original feature points in the point cloud, it exhibits relatively lower values in Precision, MCC, and IoU. PIE-Net ranks the lowest on the metrics, which is attributed to the impact of its downsampling operations. NerVE performs well on several metrics, especially Precision. EdgeFormer outperforms the compared methods in terms of Hausdorff distance, IoU, MCC and Recall. The exceptional performance of EdgeFormer is attributed to the local patch feature descriptors constructed based on the differences between feature points and non-feature points. In summary, EdgeFormer demonstrates a significant performance improvement compared to the other six methods.

% \begin{table*}[htbp]
%   \centering
%   %\fontsize{10}{12}\selectfont
%   \begin{threeparttable}
%   \caption{Comparison of different methods on the ABC dataset. The best scores are in bold}
%   \label{Table 1}
  
%     \begin{tabular}{ccccccc}
% \hline
% Method & $Hausdorff$ & $IoU$ & $MCC$ & $Precision$ & $Recall$ & $OPS$  \\ \hline
% BE & 0.486 & 0.165 & 0.182 & 0.271 & 0.469 & 0.477 \\
% PBRG & 0.316 & 0.206 & 0.261 & 0.349 & 0.408 & 0.511 \\
% SGLBP & 0.252 & 0.255 & 0.275 & 0.350 & 0.543 & 0.553 \\
% EC-Net & 0.248 & 0.066 & -0.084 & 0.081 & 0.421 & 0.402 \\
% PIE-Net & 0.354 & 0.070 & 0.161 & 0.309 & 0.082 & 0.397 \\
% NerVE & 0.252 & 0.388 & 0.541 & \textbf{0.894} & 0.401 & 0.687 \\
% EdgeFormer(ours) & \textbf{0.115} & \textbf{0.839} & \textbf{0.885} & 0.890 & \textbf{0.922} & \textbf{0.919} \\ \hline
% \end{tabular}
%     \end{threeparttable}
% \end{table*}

\subsection{Qualitative Comparison}
In this section, we first present the test results on models from the ABC dataset and then evaluate the generalization capabilities of the proposed method on the PartNet dataset.

\begin{figure}[htbp]%调节图片位置，h：浮动；t：顶部；b:底部；p：当前位置
	\centering
	\includegraphics[width=1\linewidth]{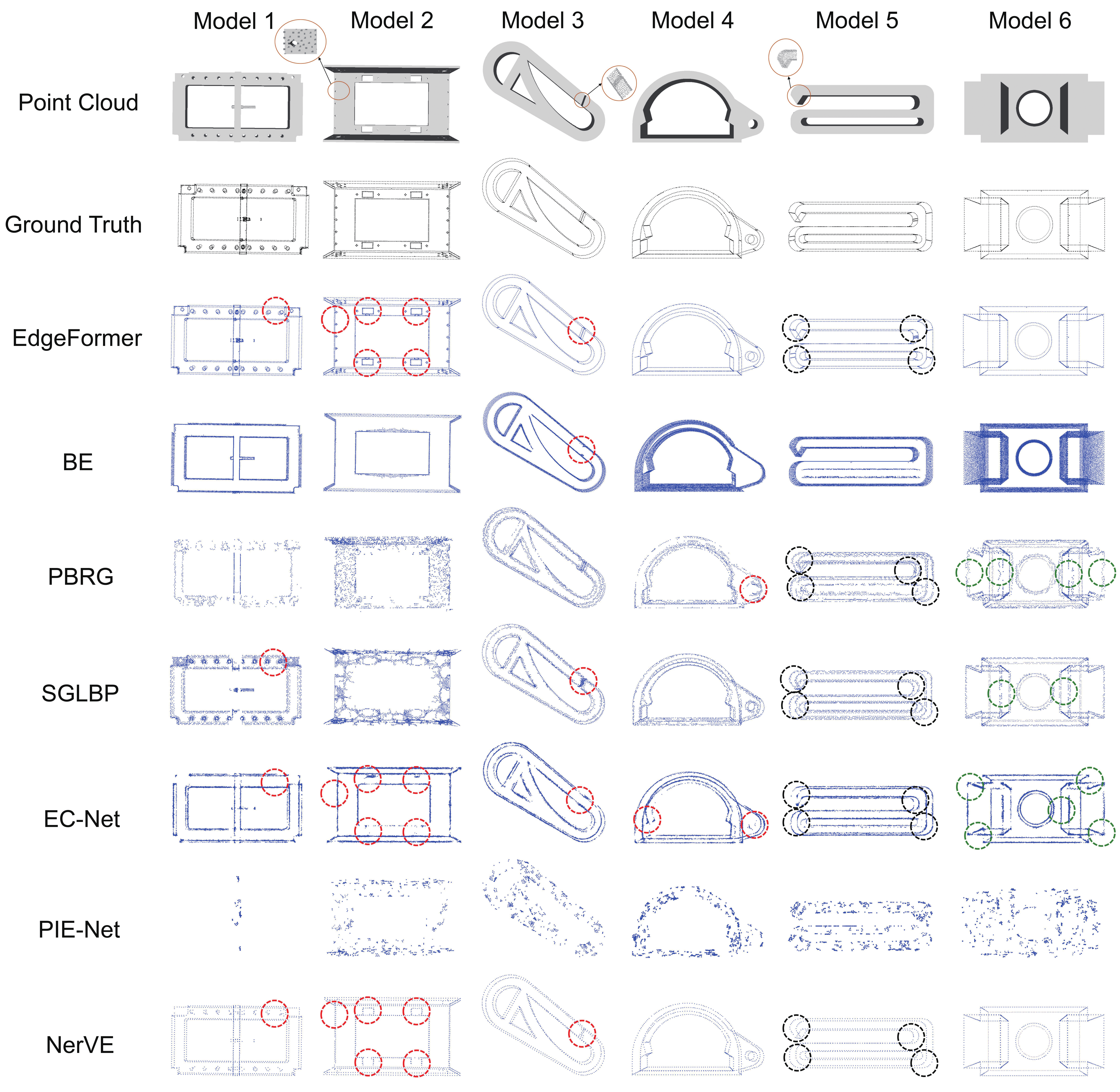}
	\caption{Edge detection results of the proposed EdgeFormer and baseline methods. Sharp features are highlighted with green circles, smooth features with black circles, and fine-grained features with red circles.}
	\label{Figure 5}
\end{figure}

\begin{figure}[htbp]%调节图片位置，h：浮动；t：顶部；b:底部；p：当前位置
	\centering
	\includegraphics[width=1\linewidth]{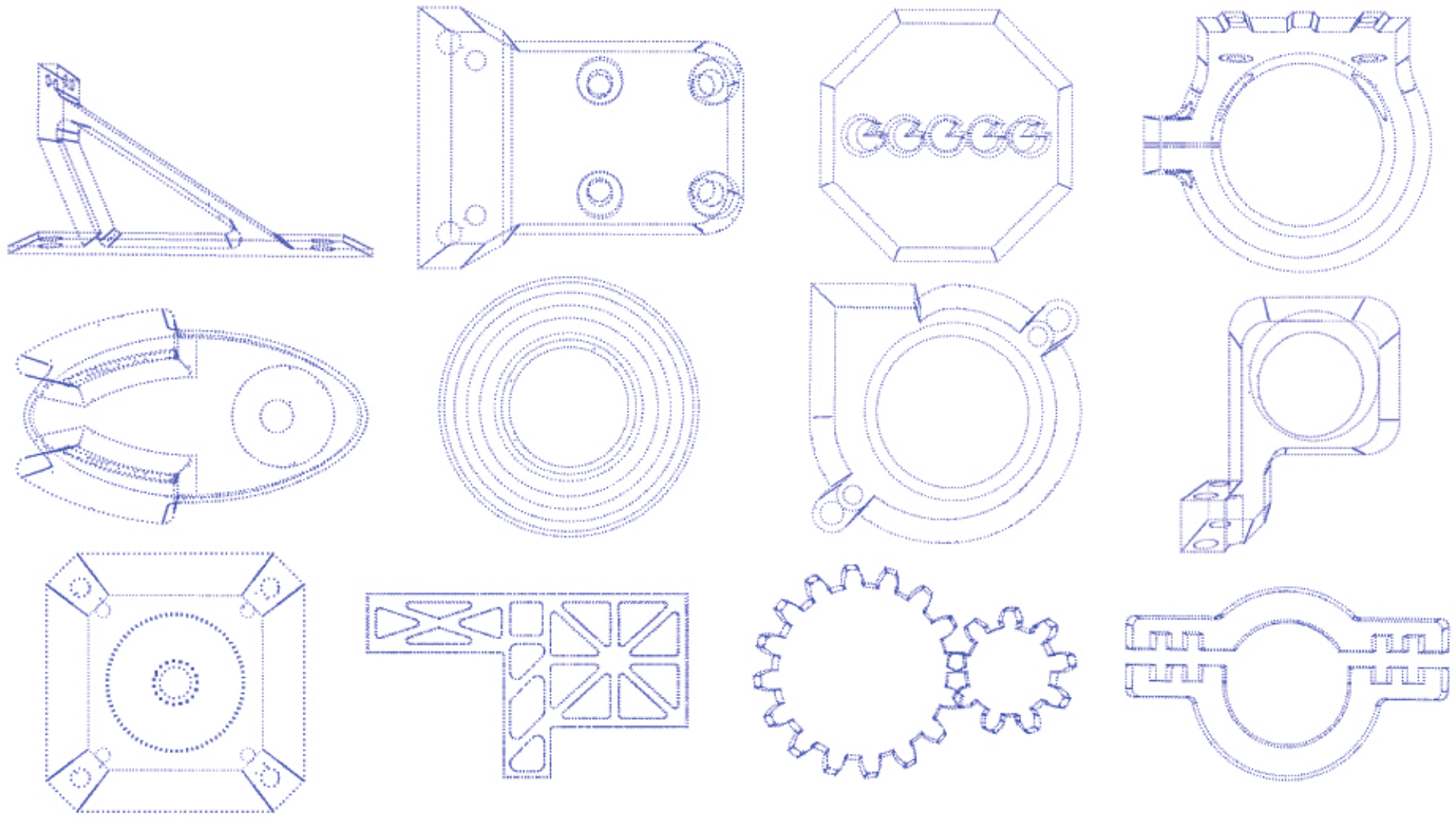}
	\caption{Qualitative results of EdgeFormer on ABC dataset}
	\label{Figure 6}
\end{figure}

Figure~\ref{Figure 5} depicts the test results of EdgeFormer on models from the ABC dataset, showcasing various shape features such as lines, circles, and B-spline curves. Particularly noteworthy are models 1, 2, and 5, which contain tiny circular and linear features, posing challenges for edge detection. EdgeFormer detects sharp, smooth, and fine-grained features, especially in models 1, 2, and 5, where it accurately identifies tiny circles and lines. It also can be found that the points detected by EdgeFormer are closer to the Ground Truth. It is important to note that the BE method has issues misidentifying non-feature points as feature points on some surfaces. While PBRG and SGLBP methods can identify most edge features, they slightly lack in capturing local details of certain models, for example, models 1 and 2 in Figure~\ref{Figure 5}. EC-Net accurately describes point cloud features but struggles with detecting some corner points. PIE-Net faces difficulties in accurately detecting most features, primarily due to the point cloud being downsampled to only 8096 points. Figure~\ref{Figure 6} displays additional edge detection results of EdgeFormer on the ABC dataset.

\begin{figure}[htbp]%调节图片位置，h：浮动；t：顶部；b:底部；p：当前位置
	\centering
	\includegraphics[width=1\linewidth]{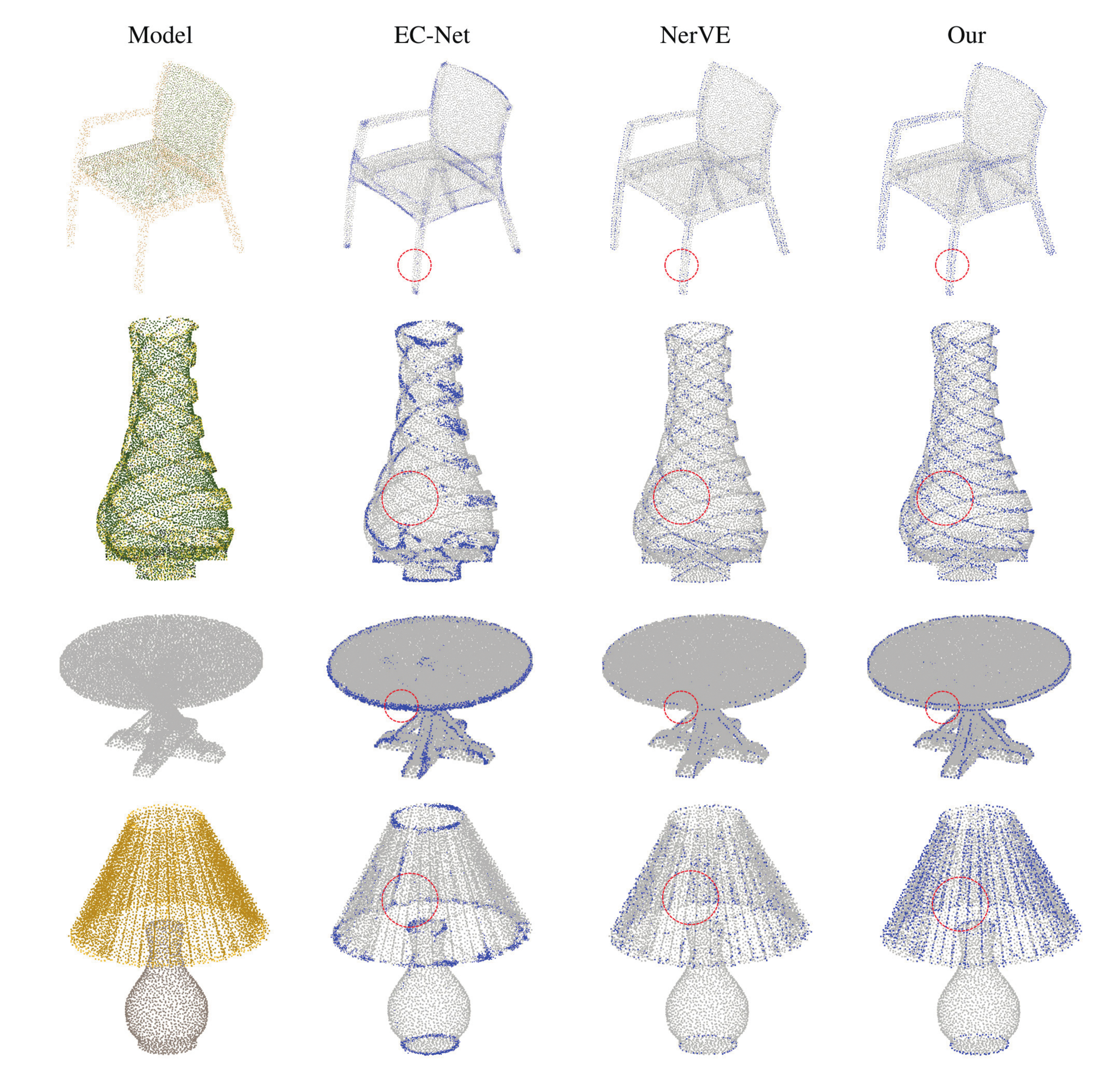}
	\caption{The results of the proposed EdgeFormer, EC-Net\cite{Yu2018ECNet}, and NerVE\cite{Zhu_2023_NerVE} on PartNet dataset.}
	\label{Figure 7}
\end{figure}

To further evaluate the generalization capabilities of EdgeFormer, we compared its performance on PartNet \cite{Mo_2019_PartNet} against two prominent deep learning methods: NerVE \cite{Zhu_2023_NerVE} and EC-Net \cite{Yu2018ECNet}. As shown in Figure~\ref{Figure 7}, the results indicate that EdgeFormer accurately detects fine-grained edges, maintaining edge integrity and avoiding the erroneous identification of non-edges. In contrast, while EC-Net and NerVE can still extract sharp features, they tend to miss finer details.

\subsection{Robustness}
We further evaluate the robustness of our model to sampling density and noise perturbation on the ABC dataset.

\textbf{Robustness to the sampling density.} To test the robustness of our method to sampling density, we use random sampling implemented in PCL to simplify the point cloud. Three scales are set: $0.8 \times N$, $0.7 \times N$, and $0.6 \times N$. $N$ is the size of the point cloud.

\begin{figure}[htbp]%调节图片位置，h：浮动；t：顶部；b:底部；p：当前位置
	\centering
	\includegraphics[width=1\linewidth]{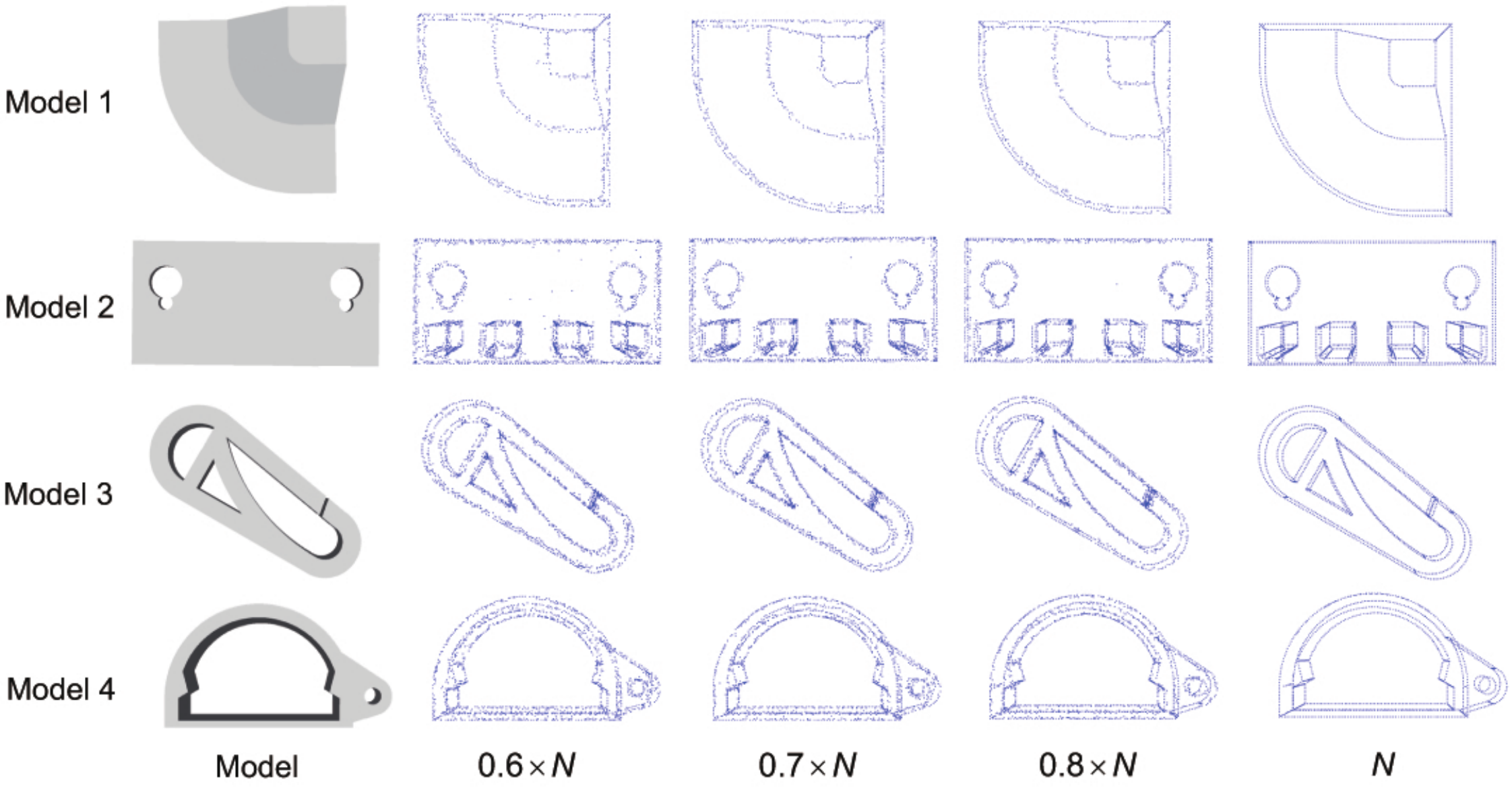}
	\caption{Edge detection results of EdgeFormer with different sampling densities}
	\label{Figure 8}
\end{figure}

The results are presented in  Figure~\ref{Figure 8}. It can be found that the proposed method can deal with models with different sampling densities; however, when the imbalance of the sampling density increases, for example, when the sampling density equals $0.6 \times N$, some points are erroneously detected.

The previous metrics highly depend on point number, we therefore use the Chamfer distance (CD)\cite{Loizou2020Learning} to qualitatively evaluate the performance on simplified models. CD is a symmetric metric that measures the difference between two point clouds and calculates the average of the nearest point distances for each point from one point cloud to the other, and the reverse average of the nearest point distances from the second point cloud to the first. This metric takes into account both overall differences in shape and local differences in detail.

\begin{equation}
    \label{Equation 18}
    CD(A,B) = \frac{1}{{|A|}}\sum\limits_{a \in A} {{{\min }_{b \in B}}} ||a - b|{|_2} + \frac{1}{{|B|}}\sum\limits_{b \in B} {{{\min }_{a \in A}}} ||b - a|{|_2}
\end{equation}
where $A$ and $B$ are two point clouds to be compared, $a_i$ are points in point cloud $A$ and $b_i$ are points in point cloud $B$.

Table~\ref{Table 2} shows the CD of EdgeFormer for edge detection on point clouds with different sampling densities. The smaller the CD value, the smaller the difference between the predicted feature point cloud and the Ground Truth, which indicates the better the edge detection. The results listed in Table~\ref{Table 2} demonstrate that, compared to the original point cloud, even when downsampled to $0.6 \times N$, the model's CD value shows slight changes, indicating that EdgeFormer maintains a degree of robustness in processing point clouds with varying densities.

\begin{table*}[htbp]
  \centering
  %\fontsize{10}{12}\selectfont
  \begin{threeparttable}
  \caption{CD of EdgeFormer on models downsampled using different simplification ratio}
  \label{Table 2}
    \begin{tabular}{ccccc}
\hline
Model Name & $0.6 \times N$ & $0.7 \times N$ & $0.8 \times N$  & $N$ \\ \hline
M1 & 0.404 & 0.354 & 0.296 & 0.176 \\
M2 & 0.620 & 0.493 & 0.395 & 0.047 \\
M3 & 1.01 & 0.802 & 0.615 & 0.014 \\
M4 & 0.901 & 0.718 & 0.534 & 0.021 \\ \hline
\end{tabular}
    \end{threeparttable}
\end{table*}

\begin{figure}[htbp]%调节图片位置，h：浮动；t：顶部；b:底部；p：当前位置
    \centering
    \includegraphics[width=1\linewidth]{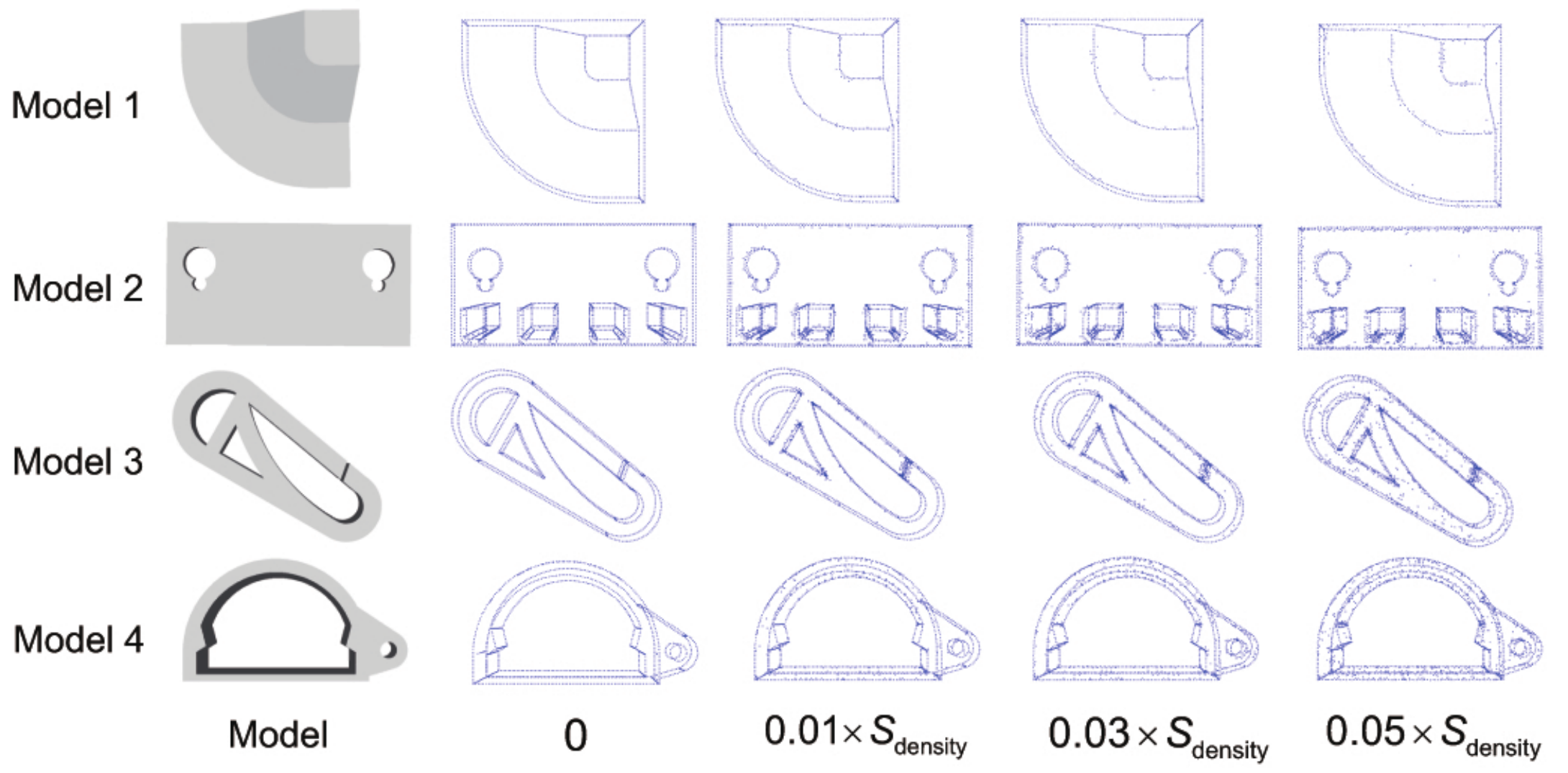}
    \caption{Edge detection results of EdgeFormer with different noise perturbations}
    \label{Figure 9}
\end{figure}

\textbf{Robustness to noises.} To test the robustness of our method to noise, we evaluate the performance of EdgeFormer on point clouds disturbed by different levels of noise. Based on the point cloud sampling density ${S_{density}}$, we set different scales of Gaussian noise, including $0.01 \times {S_{density}}$, $0.03 \times {S_{density}}$, and $0.05 \times {S_{density}}$,  which 
is formulated as Equation~\eqref{Equation 19}. The edge detection results of EdgeFormer under different noise perturbations are shown in Figure~\ref{Figure 9}. It can be seen that the method is still able to accurately detect the edge features of the point cloud and maintain high accuracy even under noise perturbations.

\begin{equation}
    \label{Equation 19}
    {S_{density}} = \frac{1}{n}\sum\limits_{i = 0}^n {di{s_i}}
\end{equation}
where 
\begin{equation}
    \label{Equation 20}
    di{s_i} = \frac{1}{k}\sum\limits_{{s_j} \in {N_i}} {||{s_i} - {s_j}|{|_2}}, 0 \le i \le n
\end{equation}
where random point set $S = \{ {s_i}\}$, ${s_i}$ is randomly selected from the point cloud, ${N_i}$ is the $k$-nearest neighbors of ${s_i}$, and ${s_j}$ is located in ${N_i}$. $|| \bullet |{|_2}$ is the Euclidean distance and we empirically set $n = \left\lfloor {\frac{N}{{10}}} \right\rfloor$ and $k$= 10.

\subsection{Ablation Study}
In this section, we conduct a comprehensive ablation study using the ABC dataset as an example. We aim to analyze the impact of the MLP block, Transformer encoder block, and local patch feature descriptors ($d_{p_ip_j}$ and $d_{p_jp_i}$) in EdgeFormer by individually omitting them from the full EdgeFormer architecture. The MLP block is utilized in the feature decoder and point classification module, while the Transformer encoder block is employed in the local patch feature enhancement module. The local patch feature descriptors serve as the input to EdgeFormer. To compare with the proposed local patch feature descriptor, classic 3D descriptors such as Spin Image\cite{spin_image}, 3DSC\cite{3dsc_frome}, SHOT\cite{shot_tombari} and PFH\cite{PFH_rusu}, implemented in PCL, were extracted from point clouds and subsequently fed into our network to evaluate their feature detection performance.

The experimental results are shown in Table~\ref{Table 3}, which presents the edge detection results of the three network architectures using only MLPs (Test \#1), only Transformer encoders(Test \#2), removing either $d_{p_ip_j}$(Test \#3) or $d_{p_jp_i}$(Test \#4), using the complete EdgeFormer architecture(Test \#5) and replacing our descriptor with the classic 3D descriptors (Test \#6-9). The scores in Table~\ref{Table 3} demonstrate that the method using the complete EdgeFormer architecture outperforms the method using only a single component in all metrics, verifying the necessity of MLPs and Transformer encoders and their combined performance. Furthermore, the ablation studies on the local patch feature descriptors $d_{p_ip_j}$ and $d_{p_jp_i}$ demonstrate that removing $d_{p_ip_j}$ slightly decreases edge detection performance, whereas removing $d_{p_jp_i}$ leads to a significant decline. Therefore, the network achieves optimal performance only when both $d_{p_ip_j}$ and $d_{p_jp_i}$ are included as inputs. Furthermore, when employing classic 3D descriptors, a significant performance gap in feature extraction emerges compared to our proposed descriptor, underscoring the superior efficacy of our local patch feature descriptor in this task.

\begin{table*}[htbp]
  \centering
  %\fontsize{10}{12}\selectfont
  \begin{threeparttable}
  \caption{EdgeFormer ablation study results on the ABC dataset. The best scores are in bold}
  \tiny
  \label{Table 3}
  \setlength{\tabcolsep}{1.2mm}
    \begin{tabular}{lcccccccc}
\hline

Method & MLPs & \multicolumn{1}{c}{\begin{tabular}[c]{@{}c@{}}Transformer\\ Encoder\end{tabular}} & $Hausdorff \downarrow$ & $IoU \uparrow$ & $MCC \uparrow$ & $Precision \uparrow$ & $Recall \uparrow$ \\ 

\hline

Test \#1 & $\surd$ &  & 0.149 & 0.754 & 0.821 & 0.826 & 0.885 \\

Test \#2 & & $\surd$ & 0.151 & 0.733 & 0.797 & 0.789 & 0.897\\
% \hline

Test \#3(w/o $D_{p_ip_j}$) & $\surd$ & $\surd$ & 0.198 & 0.709 & 0.782 & 0.760 & 0.908 \\

Test \#4(w/o $D_{p_jp_i}$) & $\surd$ & $\surd$ & 0.241 & 0.324 & 0.401 & 0.446 & 0.533\\

Test \#5 (ours) & $\surd$ & $\surd$ & \textbf{0.115} & \textbf{0.839} & \textbf{0.885} & 
\textbf{0.890} & \textbf{0.922}\\ 

\hline

Test \#6(Spin Image) & $\surd$ & $\surd$ & 0.242 & 0.453 & 0.537 & 0.549 & 0.721\\

Test \#7(3DSC) & $\surd$ & $\surd$ & 0.255 & 0.328 & 0.420 & 0.389 & 0.694\\

Test \#8(SHOT) &  $\surd$ &  $\surd$ &  0.213 &  0.549 &  0.640 &  0.626 &  0.793\\

 Test \#9(PFH) &  $\surd$ &  $\surd$ &  0.168 &  0.546 &  0.655 &  0.592 &  0.858\\

\hline

\end{tabular}
    \end{threeparttable}
\end{table*}

\subsection{Efficiency}
We selected six models in the ABC dataset to test the efficiency of the proposed method. The runtime of our method includes the local patch encoding stage and the network stage. To ensure the accuracy and reliability of the measurements, the running time of each stage was derived based on the average of three measurements. 
Table~\ref{Table 4} shows the calculation times for the two stages. It can be seen that both stages are computationally efficient, taking only 13.574 seconds to process a point cloud containing over ninety thousand points.

\begin{table*}[htbp]
  \centering
  \begin{threeparttable}
  \caption{The runtime of the proposed method.}
  \label{Table 4}
    \begin{tabular}{ccccc}
    \toprule
    \multirow{2.5}{*}{Model} & \multirow{2.5}{*}{Point Number} & \multicolumn{3}{c}{Running Time(s)} \\
    \cmidrule{3-5}
    & & Local Patch Encoding & Network & Total \\
    \midrule
    M1 & 1212  & 0.002 & 1.944 & 1.946 \\
    M2 & 5673  & 0.016 & 2.336 & 2.352 \\
    M3 & 16332 & 0.049 & 3.054 & 3.103 \\
    M4 & 22234 & 0.071 & 3.384 & 3.455 \\
    M5 & 45589 & 0.159 & 5.395 & 5.554 \\
    M6 & 97773 & 0.337 & 13.237 & 13.574 \\ \hline
\end{tabular}
    \end{threeparttable}
\end{table*}

% \begin{table*}[htbp]
%   \centering
%   %\fontsize{10}{12}\selectfont
%   \begin{threeparttable}
%   \caption{The scales of training datasets and parameter specifications between competitors and ours.}
%   \label{Table 5}
%    \color{red}
%     \begin{tabular}{ccccc}
% \hline
% Methods& \#Training Model& \#Parameters (M)\\ 
% \hline
% EC-Net\cite{Yu2018ECNet} & 36 &0.82  \\
% PIENet\cite{Wang2020PieNet}& 32& 1.03  \\
% NerVE\cite{Zhu_2023_NerVE}&1892& 4.08\\
% EdgeFormer(Ours)&50&5.65\\

% \hline
% \end{tabular}
%     \end{threeparttable}
% \end{table*}

\subsection{Limitations}
Since EdgeFormer classifies points on the point cloud into features and non-features, it may encounter failure when the point cloud is excessively sparse, leading to information loss within the local patches.

\section{Conclusions and Future Work}
The task of learning-based 3D point cloud edge detection has been faced with challenges such as difficulty in extracting fine-grained details. This paper proposes a novel network EdgeFormer for 3D point cloud edge detection. We introduce a new set of per-point features that can describe the local information of the points, referred to as the distance feature in formation in local patches. Furthermore, the EdgeFormer network is employed for feature analysis and point classification. By leveraging this network, our method achieves improved edge feature extraction, bringing it closer to the Ground Truth and enhancing the overall quality of edge detection.

In future research, we intend to explore how surface descriptors can be utilized to achieve other tasks in point clouds, such as reconstruction, classification, and segmentation.

\backmatter

\bibliography{sn-bibliography}
\end{document}